\documentclass[sigconf]{acmart}
 
\usepackage{amssymb}
\usepackage{makecell}
\usepackage{multirow}
\usepackage{balance}

\copyrightyear{2021} 
\acmYear{2021} 
\setcopyright{acmcopyright}\acmConference[MM '21]{Proceedings of the 29th ACM International Conference on Multimedia}{October 20--24, 2021}{Virtual Event, China}
\acmBooktitle{Proceedings of the 29th ACM International Conference on Multimedia (MM '21), October 20--24, 2021, Virtual Event, China}
\acmPrice{15.00}
\acmDOI{10.1145/3474085.3475508}
\acmISBN{978-1-4503-8651-7/21/10}

\AtBeginDocument{%
  \providecommand\BibTeX{{%
    \normalfont B\kern-0.5em{\scshape i\kern-0.25em b}\kern-0.8em\TeX}}}

\begin{document}
\fancyhead{}
\title{Spatiotemporal Inconsistency Learning for DeepFake Video Detection}


\author{Zhihao Gu}
\authornote{indicates equal contributions. This work was done when Zhihao Gu was a research intern at Tencent Youtu Lab.}
\orcid{1234-5678-9012}
\affiliation{%
  \institution{Shanghai Jiao Tong University}
  \streetaddress{P.O. Box 1212}
   \city{}
  \country{}
  \postcode{43017-6221}}
\email{ellery-holmes@sjtu.edu.cn}

\author{Yang Chen}
\authornotemark[1]
\affiliation{%
  \institution{Tencent Youtu Lab}
  \city{}
  \country{}}
\email{wizyangchen@tencent.com}

\author{Taiping Yao}
\authornotemark[1]
\affiliation{%
  \institution{Tencent Youtu Lab}
  \city{}
  \country{}}
\email{taipingyao@tencent.com}

\author{Shouhong Ding}
\authornotemark[2]
\affiliation{%
 \institution{Tencent Youtu Lab}
 \city{}
 \country{}}
\email{ericshding@tencent.com}

\author{Jilin Li}
\affiliation{%
 \institution{Tencent Youtu Lab}
 \city{}
 \country{}}
\email{jerolinli@tencent.com}

\author{Feiyue Huang}
\affiliation{%
 \institution{Tencent Youtu Lab}
 \city{}
 \country{}}
\email{garyhuang@tencent.com}

\author{Lizhuang Ma}
\authornote{Corresponding authors.}
\affiliation{%
 \institution{Shanghai Jiao Tong University}
 \city{}
 \country{}}
\email{ma-lz@cs.sjtu.edu.cn}

\renewcommand{\shortauthors}{Trovato and Tobin, et al.}

\begin{abstract}
  The rapid development of facial manipulation techniques has aroused public concerns in recent years. Following the success of deep learning, 
  existing methods always formulate DeepFake video detection as a binary classification problem and develop frame-based and video-based solutions. However, little attention
  has been paid to 
  capturing the spatial-temporal inconsistency in forged videos. 
  To address this issue, 
  we term this task as a Spatial-Temporal Inconsistency Learning (STIL) process and instantiate it into a novel STIL block,
   which consists of a Spatial Inconsistency Module (SIM), a Temporal Inconsistency Module (TIM), and an Information Supplement Module (ISM). 
  Specifically, we present a novel temporal modeling paradigm in TIM by exploiting the temporal difference over adjacent frames along with both horizontal and vertical directions. 
  And the ISM  simultaneously utilizes the spatial information from SIM and temporal information from TIM to establish a more comprehensive spatial-temporal 
  representation. 
  Moreover, 
  our STIL block is flexible and could be plugged into existing 2D CNNs. 
  Extensive experiments and visualizations are presented to demonstrate the eﬀectiveness of our
  method against the state-of-the-art competitors.
\end{abstract}

\ccsdesc[500]{Computing methodologies~Computer vision}

\keywords{deepfake video detection, spatiotemporal inconsistency modeling, video analysis}


\maketitle
\section{Introduction}
\begin{figure}[t!]
  \centering
  \includegraphics[width=8.5cm]{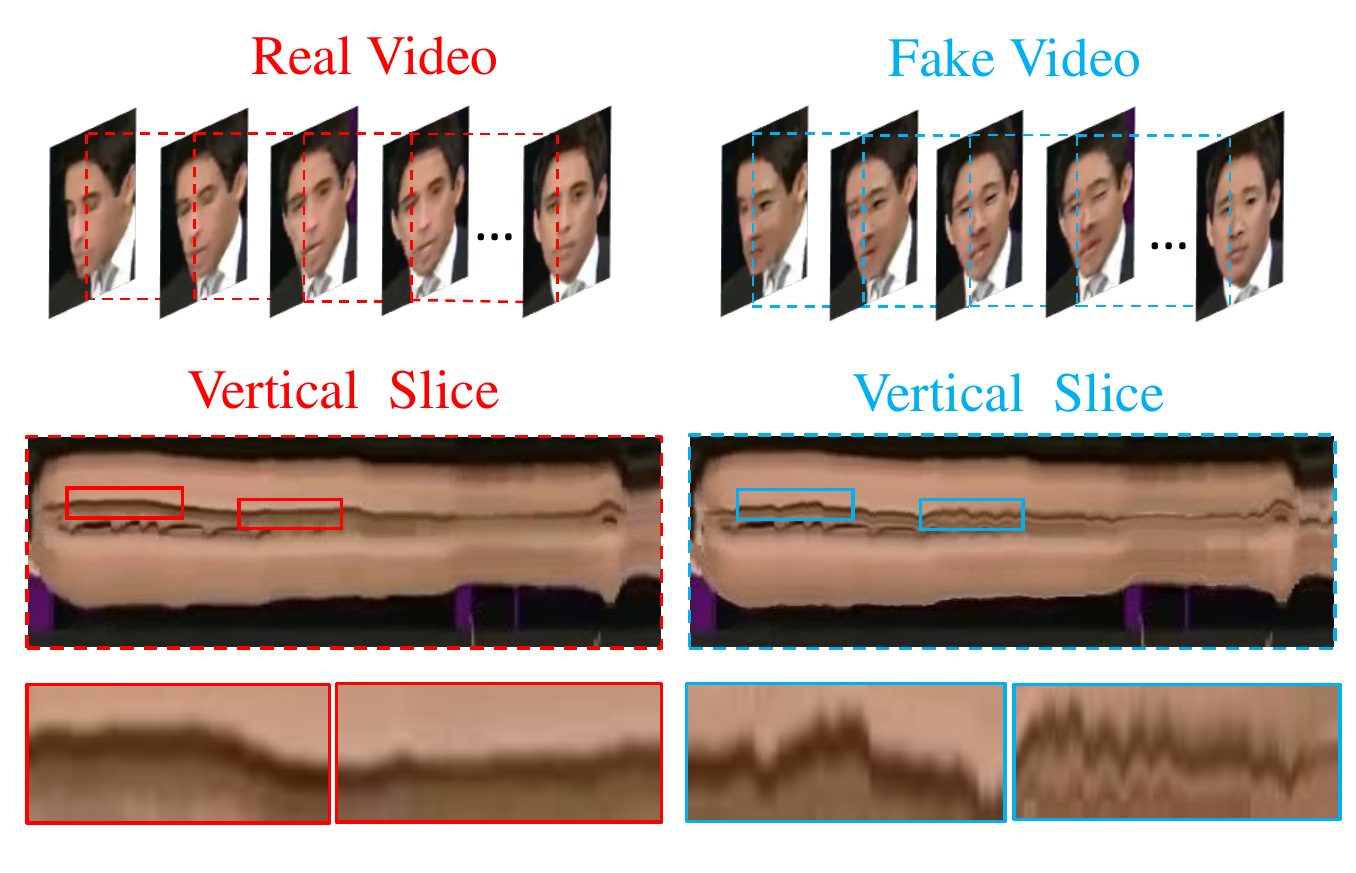}
  \caption{
  Illustration of the temporal inconsistency between real and fake videos. We visualize the motion of the vertical slice at a certain horizontal location in the video. It is obvious that the motion in the real video is smoother than the fake one, which indicates temporal inconsistency may serve as an efficient indicator for DeepFake video detection.}\label{motivation}
\end{figure}
With the explosive progress of face forgery methods in recent years, the abuse of this technology has aroused wide publicly concerns. Swapped or partially forged faces can be easily produced by publicly accessible tools, \emph{e.g.}, DeepFakes, FaceSwap, causing severe threats to cyber, social and even political security. Therefore, it's in crucial need to develop effective face forgery detection technologies.
As a binary classification problem, existing methods can be divided into frame-based and video-based methods. The frame-based methods mainly focus on mining the forgery patterns in each image, utilizing RGB information~\cite{chen2021local, WangYDM20}, frequency statistics~\cite{qian2020thinking} and auxiliary mask~\cite{dang2020detection} or blending boundary information~\cite{li2020face} for better performance. Since the forged video is generated through frame-by-frame manipulation, there often exists visually unnatural image transition across frames or temporal inconsistency like facial position jittering. Advanced forgery techniques may create extremely genuine facial images but cannot eliminate this temporal inconsistency. Therefore, the frame-based methods fail to catch this temporal discriminative feature and have limited performance. Many researchers recently develop video-based methods in face forgery detection. General video analysis models like C3D \cite{tran2014c3d}, I3D \cite{carreira2017quo} and LSTM \cite{hochreiter1997long} are applied in this area. Besides the high computational cost, these methods are not specifically designed for face forgery detection and have limited capability of catching the temporal inconsistency. And their performances are even not comparable to frame-based methods by averaging frame-level scores to obtain the video-level decision.

\begin{figure*}[ht]
  \centering
  \includegraphics[height=9cm, width=\linewidth]{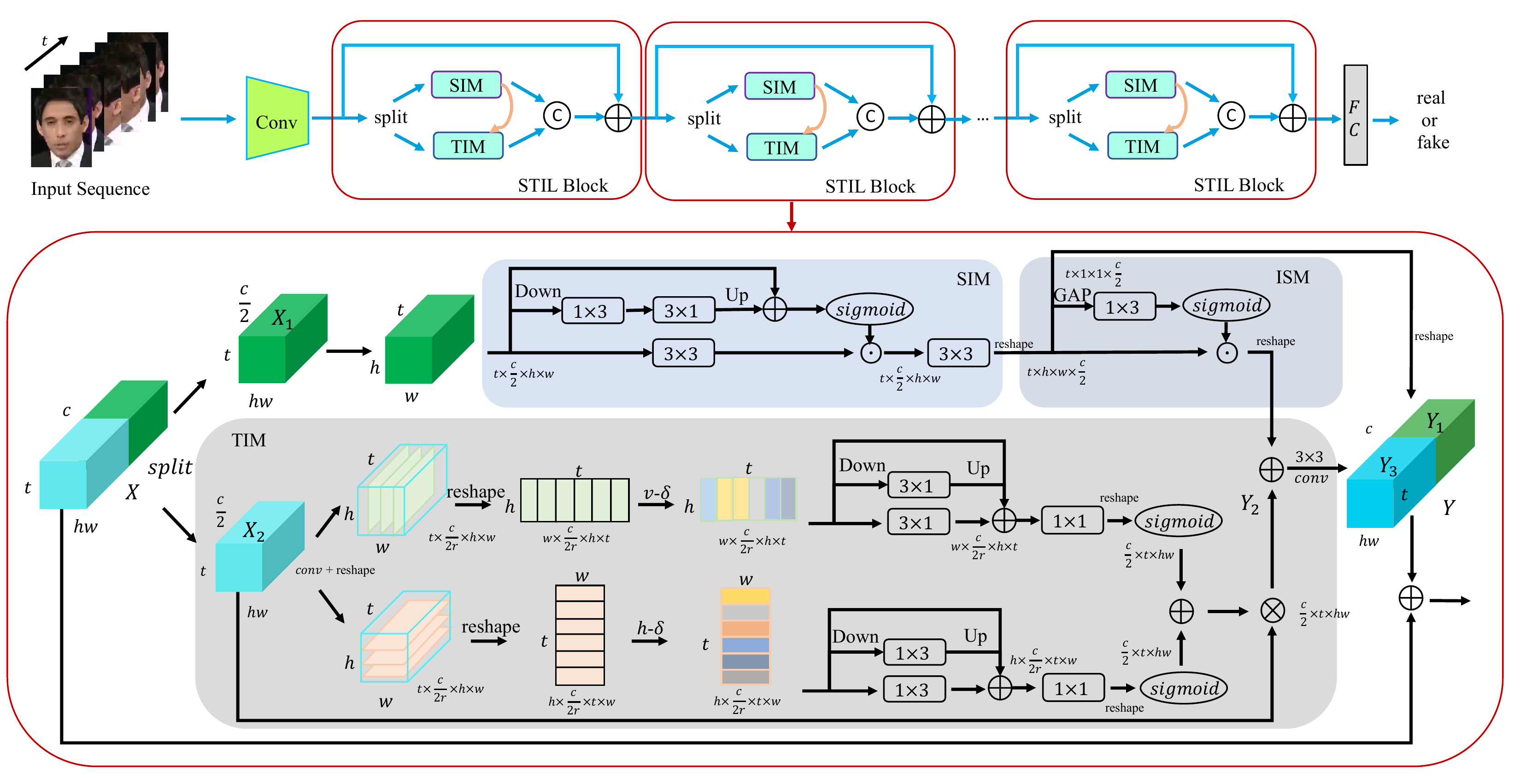}
  \caption{The structure of the proposed Spatial-Temporal Inconsistency Learning (STIL) block. Each STIL block consists of a Spatial Inconsistency Module (SIM), a Temporal Inconsistency Module, (TIM) and an Information Supplement Module (ISM). The SIM aims to capture the spatial inconsistency within each frame, \iffalse by 2D CNNs \fi and TIM applies two orthogonal temporal difference operations over adjacent frames to capture the temporal inconsistency. ISM builds information flow from SIM to TIM for a more comprehensive feature representation. \iffalse The STIL block is specially designed for face forgery detection and could be plugged into existing 2D CNNs.\fi}\label{framework}
  \Description{A woman and a girl in white dresses sit in an open car.}
\end{figure*}

In view of the previous works, we argue that it's necessary to take both the spatial and temporal information into consideration in face forgery detection. The trace that distinguishes between real and forged videos can be unified concluded as a form of inconsistency, which not only exists within each image but also across frames along the temporal dimension. Motivated by \cite{liu2020teinet} which utilizes the temporal difference in action recognition, we develop a new modeling paradigm in face forgery detection, termed as Temporal Inconsistency Module (TIM), to capture the temporal inconsistency effectively. As shown in Fig~\ref{motivation}, we slice each frame vertically at a fixed horizontal location in a video and then concatenate them to form a $h-t$ map, the patterns of the real one stretch smoothly whereas the forged one depicts discontinuous burrs. Similar observation can also be obtained in the $w-t$ map. This difference demonstrates the temporal inconsistency of facial areas in a novel perspective. 
Concretely, our proposed TIM first slices the feature map along with horizontal and vertical directions and then exploits the temporal difference over adjacent frames as a guide to focus on inconsistent areas. Since TIM explicitly models the temporal inconsistency, this architecture can effectively locate the temporally forged trace in videos.

Moreover, we propose a novel and flexible block, referred as to Spatiotemporal Inconsistency Learning (STIL) block, to integrate both spatial and temporal features in a unified 2D CNN framework. In our design, the STIL block works in a two-stream manner, the first of which is termed as Spatial Inconsistency Module (SIM) and aims to explore the spatial forged patterns or spatial inconsistency, and the second of which is TIM for temporal inconsistency mining. Besides, we further propose an Information Supplement Module (ISM) at the end of the STIL block. This module fuses and enhances the two-stream spatial-temporal features for a more comprehensive facial trace representation. Our proposed STIL block is specially designed for face forgery detection and could be plugged into existing 2D CNNs like ResNets. The main contributions are as follows:
\begin{itemize}
\item We formulate DeepFake video detection as a spatial and temporal inconsistency learning process and instantiate it into our proposed STIL block, which effectively utilizes both spatial and temporal information for a more comprehensive representation. Besides, the STIL block can be plugged into existing 2D CNNs.

\item We present a novel temporal inconsistency modeling paradigm, termed as TIM, which utilizes the temporal difference between adjacent frames along with both horizontal and vertical directions. TIM works effectively by explicitly modeling the temporal inconsistency within forged videos.

\item Extensive experiments and visualizations are presented to demonstrate the effectiveness of our model, which outperforms the state-of-the-art methods on four widely used public benchmarks.
\end{itemize}

\section{RELATED WORK}
\subsection{DeepFake Generation}

Early researchers develop hand-crafted features for DeepFake generation. These methods typically swap the facial areas based on face landmarks and utilize post-processing techniques to make the forged boundaries inconspicuous. \cite{dale2011video} uses a 3D multilinear model to track the facial performance in videos to minimize the blending trace. \cite{garrido2014automatic} proposes a facial reenactment system to replace the target face while preserving the original performance. Since these traditional methods have limited face forgery effects, researchers begin to employ deep learning techniques for more realistic face generation. \cite{thies2019deferred} introduces Deferred Neural Rendering for image synthesis through combining traditional graphics pipelines with learnable elements which is trained with U-Net~\cite{ronneberger2015u} in an end-to-end manner. ~\cite{li2019faceshifter} proposes a two-stage network to first synthesize high-fidelity swapped faces and then recover anomaly regions in a self-supervised way. Despite the high-quality forgery effects of these methods, they basically works frame-by-frame and the forgery trace across frames is hard to eliminate, which can still serves as an important discriminative clue.



\subsection{DeepFake Detection}
The abuse of DeepFake generation has caused severe security issues in recent years. To this end, lots of methods are proposed to identify these fake faces. Early researches \cite{fridrich2012rich,pan2012exposing} focus on designing hand-crafted features for DeepFake detection as face forgery techniques are limited at that time. With the development of deep learning, the forgery faces are more and more realistic. Therefore, some works utilize state-of-the-art DNNs to extract discriminative frame-level features for DeepFake detection. \cite{rossler2019faceforensics++} uses the effective XceptionNet to identify DeepFakes. \cite{dang2020detection} proposes to utilize an attention mechanism to process and improve the feature maps. \cite{li2020face} presents the face X-Ray which checks whether an image can be decomposed into the blending of two images. All these methods achieve high accuracy for image-level DeepFake detection. However, with the development of DeepFake generators, these image-based detectors may fail to capture spatial and temporal inconsistency across multiple frames.
Recent works consider DeepFake detection as a video-level prediction problem and learn video-level features. \cite{sabir2019recurrent} uses recurrent convolutional networks to exploit the temporal information from image streams across domains. \cite{li2020sharp} presents the S-MIL-T, which uses Xecption-Net to extract instance embeddings and temporal CNNs for bag predictions. \cite{zi2020wilddeepfake} leverages the attention masks to enhance the backbone representations. \cite{qi2020deeprhythm} proposes to monitor the heartbeat rhythms to expose the DeepFakes. However, they explicitly follow the paradigm that either combining well studied 2D backbones with temporal operations, \emph{e.g.}, recurrent neural network (RNN) or averaging the frame-level representations, whose performance is largely effected by the representation capacity of 2D backbones.
Different from existing methods, we rethink the characters of DeepFake detection and formulate it as a spatial and temporal inconsistency learning process, and accordingly, we design a STIL block to model the inconsistency.

\subsection{Video Temporal Modeling}
Temporal modeling is essential for video-related tasks. Many researches~\cite{tran2014c3d,carreira2017quo,lin2019temporal,hochreiter1997long,liu2020teinet,wang2020tdn} have been devoted to designing effective architectures for capturing motion information. \cite{tran2014c3d} proposes a simple, yet effective structure using 3-dimensional CNNs for spatiotemporal feature learning. \cite{carreira2017quo} introduces a Two-Stream Inflated 3D CNN to leverage ImageNet pre-trained architecture and seamlessly learn spatiotemporal features.
Since these 3D-convolution-based methods contain an order of magnitude more parameters than 2D counterparts, they are computationally expensive to deploy and thus efficiency becomes a big challenge. To alleviate this, a lot of efficient modelings are proposed, enabling 2D CNNs to model the temporal information. \cite{lin2019temporal} proposes to shift part of the channels along the temporal dimension, facilitating to exchange information among adjacent frames. \cite{liu2020teinet} decouples the channel correlation and temporal interaction to learn temporal features. \cite{wang2020tdn} leverages the temporal difference operator on neighboring frames. These methods outperform their 3D counterparts while achieving high efficiency. 
Different from these methods, our STIL block is specifically designed for DeepFake video detection and  utilize temporal difference within adjacent frames along two orthogonal directions to capture fine-grained temporal inconsistency, which is more effective in this task.

\section{Overview}
We treat the DeepFake video detection as a binary classification problem. Given an input sequence with the shape of $T \times C \times H \times W$, where $T, C, H, W$ denote the input frame number, image channel number, image height, and width separately, the 2D CNN model outputs the video-level prediction to decide whether this video is forged or not. We formulate this problem as a spatiotemporal inconsistency learning process and instantiate it into the Spatiotemporal Inconsistency Learning (STIL) block. The STIL block is specially designed for DeepFake video detection and Fig~\ref{framework} depicts how it works when plugged into the ResNet block.

The STIL block works in a two-stream manner and consists of three modules: Spatial Inconsistency Module (SIM), Temporal Inconsistency Module (TIM), and Information Supplement Module (ISM). In consideration of computation efficiency, we first split the input $\boldsymbol{X}$ along the channel dimension into two portions $\{\boldsymbol{X_1}, \boldsymbol{X_2}\}$ at the entry of STIL block. Without loss of generality, the split operation is performed uniformly and each portion has half of the channel number of $\boldsymbol{X}$. These two portions are then respectively sent into two special pathways for capturing forgery inconsistency from two different perspectives. In the first pathway, we adopt the SIM upon $\boldsymbol{X_1}$ to capture the spatial forged patterns within each image, yielding $\boldsymbol{Y_1}$. In the second pathway, TIM focuses on mining the temporal inconsistencies introduced by frame-by-frame face manipulation methods from $\boldsymbol{X_2}$ and embeds this discriminative feature into output $\boldsymbol{Y_2}$. Additionally, ISM builds an information flow from the spatial pathway to the temporal pathway, aiming to explore the complementarity between them. ISM enhances $\boldsymbol{Y_2}$ by learning from $\boldsymbol{Y_1}$ and generates a more comprehensive temporal representation $\boldsymbol{Y_3}$. Finally, the two-stream intermediate outputs $\{\boldsymbol{Y_1}, \boldsymbol{Y_3} \} $ are concatenated and fused to form $\boldsymbol{Y}$, which encodes the discriminative features in both spatial and temporal dimensions. In the following sections, we will describe these three modules in detail.

\subsection{Spatial Inconsistency Module}
During the process of face forgery, the produced manipulated image is often accompanied by distinguishable artifacts in the facial area. For example, the up-sampling operation in GAN-based methods may cause the checker-board pattern~\cite{huang2020fakepolisher}. Moreover when merging the swapped facial part to the original image, the blending borders or image quality mismatch is hard to eliminate. We term these discriminative features as the spatial inconsistency, which exists between the pristine and forged areas in each image. Many frame-based works \cite{li2020face}\cite{zhao2021multi} have shown that making use of the spatial inconsistency indeed leads to promising results.

To this end, we design a Spatial Inconsistency Module (SIM) as part of the STIL block to capture the spatial inconsistency. As depicted in Fig~\ref{framework}, SIM is composed of a series of 2D operations in the image level without taking the temporal information into consideration. To achieve optimal performance,
we adopt a series of good practices in model design and propose a three-path architecture in SIM. Given input feature map $\boldsymbol{X_1}$ with shape $T \times \frac{C}{2} \times H \times W $, the middle pathway works in a low-resolution manner by applying an average pooling with kernel size $2 \times 2$ and stride $2$ upon $\boldsymbol{X_1}$, followed by two consecutive convolutions and a bilinear up-sampling operation: 
\begin{equation}
    S= {\rm Up}(K_1*K_2*({\rm AvgPool_2}(X_1))).
\end{equation}
where $K_1$ and $K_2$ are convolutions of kernel size $1\times3$ and $3\times1$, and $S$ is the output which has the same shape with $\boldsymbol{X_1}$. This step allows for a larger receptive field and leads to better performance when coupled with other pathways with normal receptive fields. By adding $X_1$ by a res-connection, the information lost in the downsampling progress is saved. Finally, the confidence scores are assigned and the spatial inconsistency areas are highlighted:
\begin{equation}
    {\rm Y_1}= K_4*(\sigma(S+X_1)\odot K_3(X_1)).
\end{equation}
where $K_3$ is a $3 \times 3$ convolution for feature extraction, $K_4$ is a $3 \times 3$ convolution for post-processing and $\sigma$ is the sigmoid function. We visualize the attention maps in Fig \ref{differentheatmap} and Fig  \ref{stbeforefinalheatmap} for illustration.

\subsection{Temporal Inconsistency Module}
Although spatial inconsistency plays a key role in identifying face forgery attacks, the forgery techniques develop rapidly and some cutting-edge methods may produce extremely genuine forged facial images. Chosen a single image from them, it is even challenging for human beings to distinguish. However, since the forged video is generated through frame-by-frame manipulation, the inter-frame temporal inconsistency like facial position jittering is always present and relatively more visually distinguishable,  which provides another new  perspective in forgery detection. 

To this end, we propose a new modeling paradigm, termed as Temporal Inconsistency Module (TIM), to explicitly model the temporal inconsistency within forged videos. We find that this temporal inconsistency trace is most salient if observed along with the horizontal and vertical directions separately, which is intuitively depicted in Fig~\ref{motivation}. Consequently, we attend the network by exploiting the feature difference over adjacent frames along these two orthogonal directions to focus on the temporal inconsistent regions. As depicted in Fig~\ref{framework}, the input $\boldsymbol{X_2}$ with shape $T \times \frac{C}{2} \times H \times W $ is fed along the horizontal and vertical pathway separately, and then followed by convolution, difference, and sigmoid operations to obtain two importance weights $F_h$ and $F_w$ with the same shape as $\boldsymbol{X_2}$.

\begin{table*}
  \caption{
  Comparison on FF++ dataset under different compression rates. $\dag$ implies re-implementation and results of other methods are directly from their original papers for fair comparison. The best results are highlighted.}
  \label{ffpp}
  \begin{tabular}{ccccccccc}
    \toprule
     \multirow{2}{*}{Methods}      & \multicolumn{4}{c}{FaceForensics++ c23}  & \multicolumn{4}{c}{FaceForensics++ c40}\\
     \cmidrule(lr){2-5}\cmidrule(lr){6-9} & DF & F2F & FS & NT & DF & F2F & FS & NT\\
    \midrule
    XN-avg \cite{rossler2019faceforensics++} & 0.9893 & 0.9893 & 0.9964 & 0.9500 & 0.9678 & 0.9107 & 0.9464 & 0.8714\\ [3pt]
    C3D \cite{tran2014c3d} & 0.9286 & 0.8857 & 0.9179 & 0.8964 & 0.8929 & 0.8286 & 0.8786 & 0.8714 \\ [3pt]
    I3D \cite{carreira2017quo} & 0.9286 & 0.9286 & 0.9643 & 0.9036 & 0.9107 & 0.8643 & 0.9143 & 0.7857\\ [3pt]
    LSTM \cite{hochreiter1997long} & 0.9964 & 0.9929 & 0.9821 & 0.9393 & 0.9643 & 0.8821 & 0.9429 & 0.8821\\ [3pt]
    TEI \cite{liu2020teinet} & 0.9786 & 0.9714 & 0.9750 & 0.9429 & 0.9500 & 0.9107 & 0.9464 & 0.9036 \\ [3pt]
    FaceNetLSTM \cite{sohrawardi2019poster} & 0.8900 & 0.8700 & 0.9000 & - & - & - & - & - \\ [3pt]
    Comotion-35 \cite{wang2020exposing} & 0.9595 & 0.8535 & 0.9360 & 0.8825 & 0.9160 & - & - & -\\ [3pt]
    Comotion-70 \cite{wang2020exposing} & 0.9910 & 0.9325 & 0.9830 & 0.9045 & - & - & - & -\\ [3pt]
    DeepRhythm \cite{qi2020deeprhythm} & 0.9870 & 0.9890 & 0.9780 & - & - & - & - & - \\ [3pt]
    ADDNet-3d \cite{zi2020wilddeepfake}$^{\dag}$ & 0.9214 & 0.8393 & 0.9250 & 0.7821 & 0.9036 & 0.7821 & 0.8000 & 0.6929 \\ [3pt]
    S-MIL-T \cite{li2020sharp} & \textbf{0.9964} & \textbf{0.9964} & \textbf{1.0000} & 0.9429 & 0.9714 & 0.9107 & 0.9607 & 0.8679 \\ [3pt]
    \midrule
     STIL (ours)  & \textbf{0.9964} & 0.9929 & \textbf{1.0000} & \textbf{0.9536} & \textbf{0.9821} & \textbf{0.9214} & \textbf{0.9714} & \textbf{0.9178} \\ [3pt]
    \bottomrule
  \end{tabular}
\end{table*}

\begin{table}[htbp]
  \centering
  \caption{
  Comparison on Celeb-DF, DFDC, and WildDeepfake datasets. Best results are highlighted.}
  \label{celeb}
    \begin{tabular}{lccc}
    \toprule
    \multicolumn{1}{c}{Methods} & Celeb & DFDC  & \multicolumn{1}{p{4em}}{Wild-DF} \\ [4pt]
    \midrule
    XN-avg \cite{rossler2019faceforensics++} & 0.9944 & 0.8458 & 0.8325 \\ [4pt]
    I3D \cite{carreira2017quo}  & 0.9923 & 0.8082 & 0.6269 \\ [4pt]
    TEI \cite{liu2020teinet}$^{\dag}$ & 0.9912 & 0.8697 & 0.8164\\ [4pt]
    LSTM \cite{hochreiter1997long} & 0.9573 & 0.7902 & - \\ [4pt]
    D-FWA \cite{li2018exposing} & 0.9858 & 0.8511 & - \\ [4pt]
    ADDNet-3D \cite{zi2020wilddeepfake}$^{\dag}$ & 0.9516 & 0.7966 & 0.655 \\ [4pt]
    S-IML-T \cite{li2020sharp} & 0.9884 & 0.8511 & - \\ [4pt]
    \midrule
    STIL (ours)  & \textbf{0.9978} & \textbf{0.8980} & \textbf{0.8412} \\ [4pt]
    \bottomrule
    \end{tabular}
\end{table}

In practice, take the vertical pathway which produces $F_h$ for example, the input $\boldsymbol{X_2}$ is first compressed in the channel dimension by a factor $r$ and reshaped to $\boldsymbol{X_2^h} \in R^{W \times \frac{C}{2r} \times H \times T} = \{ x_1^h, x_2^h, \cdots, x_T^h \} $, where $x_j^h \in R^{W \times \frac{C}{2r} \times H}, 1 \le j \le T$. Then the temporal difference calculation along $H$ direction can be formulated as:
\begin{align}
    s_t^h = {\rm{Conv1}} (x_{t+1}^h) - x_{t}^h, \ \ \ t=1,2, \dots, T-1
\end{align}
\noindent where ${\rm{Conv1}}$ is a $3 \times 1$ 2D convolution. Note that since there is no more temporal difference information for $T^{th}$ frame, we set $s_T^h$ as zero map for simplification. Then we have the vertical slice difference map $\boldsymbol{S^h} \in R^{W \times \frac{C}{2r}\times H\times T} = \{ s_1^h, s_2^h, \cdots, s_T^h \} $, and similarly, the horizontal slice difference map $\boldsymbol{S^w} \in R^{H \times \frac{C}{2r}\times T\times W} = \{ s_1^w, s_2^w, \cdots, s_T^w \} $. The temporal difference operation is simple and 
does not introduce any extra parameters but capable of modelling the temporal inconsistency efficiently.

To capture the temporal inconsistency, 
we further design a multi-fields-of-views structure for efficient inconsistency extraction, which contains two parts, \emph{i.e.}, a vertical temporal inconsistency enhancement (VTIE) for $\boldsymbol{S^h}$ and a horizontal temporal inconsistency enhancement (HTIE) for $\boldsymbol{S^w}$. Here, we just elaborate on the $\boldsymbol{S^h}$ part. As shown in the Fig \ref{framework}, it takes $\boldsymbol{S^h}$ as input and has three branches to extract multi-level representations: 1) a $3\times1$ convolution along the horizontal dimension, 2) an average pooling operation, a $3\times1$ convolution along the horizontal dimension and an up-sampling pooling operation, and 3) a skip connection. The $\boldsymbol{S^w}$ part has a similar structure except for the $1\times3$ convolution along the vertical dimension. An element-wise addition is used to fuse these features, followed by a $\sigma$ function to decide importance. Finally, the confidence map is multiplied to $X_2$ to emphasize the inconsistency along temporal dimension:
\begin{equation}
   {\rm Y_2} = \frac{1}{2}[F_h + F_w] \odot X_2 = \frac{1}{2}[{\rm VTIE}(\boldsymbol{S^h}) + {\rm HTIE}(\boldsymbol{S^w})] \odot X_2.
\end{equation}
TIM guides the network to focus on the forged trace caused by temporal inconsistency and attends the feature map by learnt importance weights along the horizontal and vertical directions. This design is not only efficient but also effective, which is studied in the ablation study.

\subsection{Information Supplement Module}
While SIM and TIM capture spatial and temporal inconsistency separately, a natural question is whether they can provide each other with complementary information. To this end, we design an Information Supplement Module (ISM) to explore it. Concretely, we build information flow from one module to the other, with the aim to facilitate the feature representation of the latter one. Here we
study three forms of the information flow: (1) unidirectional connection from SIM to TIM, denoted as $\mathcal{S}\rightarrow \mathcal{T}$; (2) unidirectional connection from TIM to SIM, denoted as $\mathcal{T}\rightarrow \mathcal{S}$; (3) bidirectional connections between TIM and SIM, denoted as 
$\mathcal{T} + \mathcal{S}$. Interestingly, $\mathcal{S}\rightarrow \mathcal{T}$ achieves the best performance in practice. Related analysis can be found in the ablation study.

Specifically, given the output representation $\boldsymbol{Y_1}$ from SIM, useful channels for supplement are selected through an undergoing combination of a global average pooling (GAP) and a 1D convolution. GAP is to obtain a global representation, and the 1D convolution is applied in the channel dimension to capture the cross-channel dependencies for further assigning the importance of each channel. After that, essential channels are thus emphasized:
\begin{equation}
  \bar{Y_1} = \sigma(K_5 * ({\rm GAP}(Y_1)))\odot Y_1.
\end{equation}
where $K_5$ is the 1D convolution with kernel size 3.
Finally, we build an information flow from $\mathcal{S}$ to $\mathcal{T}$ by fusing the two-stream features:
\begin{equation}
  {Y_3} = K_6 * (\bar{Y_1} + Y_2).
\end{equation}
where $K_6$ is a $3 \times 3$ convolution. To better understand the complementarity between them, we conduct the corresponding experiments in the ablation study.

\section{EXPERIMENTS}
In this section, we provide a systematic evaluation of our method. First, we compare it with the state-of-the-art on four widely used benchmark face forgery datasets under video-level settings. Then, we conduct experiments across different datasets to demonstrate the generalization capability. Finally, a series of ablation studies are performed to assess the impact of several key components.

\subsection{Experimental Datasets}
\textbf{FaceForensics++ (FF++)} \cite{rossler2019faceforensics++}: FF++ is a standardized benchmark for evaluation of face forgery detection methods. It consists of 1000 real videos extracted from YouTube and four manipulation techniques are used to generate corresponding fake videos, \emph{i.e.}, DeepFakes (DF), Face2Face (F2F), FaceSwap (FS), and NeuralTextures (NT), with two different compression rates.\\
\textbf{Celeb-DF} \cite{li2020celeb}: Celeb-DF is a large-scale dataset in DeepFake detection for developing and evaluating DeepFake detection algorithms, which comprises 5,639 high-quality DeepFake videos from YouTube and contains more than 2 million frames in total. The fake videos are generated by an improved synthesis method and thus the overall quality presents fewer notable visual artifacts compared to previously available datasets.\\
\textbf{Deepfake Detection Challenge (DFDC)} \cite{dolhansky2019deepfake}: DFDC dataset is a preview dataset consisting of 1,131 real videos from 66 paid actors and 4,119 fake videos which are generated by two synthesis methods of unknown kind. Different from FF++ which provides frame-level labels and all the frames are manipulated in a fake video, it only provides video-level labels and there even exists some un-manipulated frames in a fake video.\\
\textbf{WildDeepfake} \cite{zi2020wilddeepfake}: WildDeepfake is a recently proposed real-world dataset. It consists of 7,314 face sequences extracted from 707 fake videos and completely come from the internet. It can be used in addition to existing datasets for developing and evaluating DeepFake video detectors. As assessed in the paper, it is more challenging compared to previously proposed datasets.

\subsection{Implementation Details}
Following \cite{li2020sharp}, we choose Dlib as the face detector for the FF++ dataset, and MTCNN for other datasets. Frames are sampled uniformly from each video and only the facial areas are cropped out as the model input. For each video, we use 8 frames to train all frame-based and video-based models and use 16 frames during the test. We adopt ResNet50 as the basic backbone and plug our STIL block into each bottleneck block to substitute the $3\times3$ convolution. In network design, the channel compression ratio $r$ in the split operation of TIM is set to $16$. The input size is $224 \times 224$ in RGB format. We adopt Adam \cite{kingma2014adam} as the optimizer supervised by binary cross-entropy loss and train our model for 30 epochs. The batch size is $16$ and the initial learning rate is 0.0002 decayed by a factor of 10 after every 10 epochs of training. Only horizontal flip is employed during training for augmentation as we focus mainly on network design.

\subsection{Baselines}
To demonstrate the advantage of our proposed architecture, we select a series of most representative works in face forgery detection as comparison. For frame-base methods, we choose Xception~\cite{rossler2019faceforensics++} and average the frame-level scores to give the final video-level prediction, which is denoted as XN-avg for simplification. While verifying the effectiveness of classic models in action analysis, we conduct a comprehensive comparison with 3D convolution based works C3D~\cite{tran2014c3d}, I3D~\cite{carreira2017quo}, 2D-Convolution-plus-RNN works LSTM~\cite{hochreiter1997long}, and 
the latest advanced TEI~\cite{liu2020teinet} which develop efficient spatial-temporal modelling techniques upon 2DCNN. In terms of the state-of-the-art works in DeepFake video detection, we choose FaceNetLSTM~\cite{sohrawardi2019poster}, D-FWA~\cite{li2018exposing}, Comotion~\cite{wang2020exposing}, DeepRhythm~\cite{qi2020deeprhythm}, ADDNet-3D~\cite{zi2020wilddeepfake} and S-IML-T~\cite{li2020sharp}. 

\begin{table}[htbp]
  \centering
  \caption{Cross dataset generalization comparison in terms of AUC. $\dag$ implies re-implementation.}
  \label{cross-dataset}
  \setlength{\tabcolsep}{4mm}
    \begin{tabular}{lcc}
    \toprule
    Methods  & FF++ DF  & Celeb-DF\\ [4pt]
    \midrule
    HeadPose \cite{yang2019exposing} & 0.4730 & 0.5460 \\ [4pt]
    D-FWA \cite{li2018exposing}  & 0.8100  & 0.5690 \\ [4pt]
    VA-LogReg \cite{matern2019exploiting} & 0.7800  & 0.5510 \\ [4pt]
    Xception-c40 \cite{rossler2019faceforensics++} & 0.9550 & 0.6550 \\ [4pt]
    Multi-task \cite{nguyen2019multi} & 0.7630 & 0.5430 \\ [4pt]
    Capsule \cite{nguyen2019capsule} & 0.9660 & 0.5750 \\ [4pt]
    DoubleRNN \cite{masi2020two} & 0.9318 & 0.7341 \\ [4pt]
    ADDNet-3D \cite{zi2020wilddeepfake}$^{\dag}$ & 0.9622 & 0.6085\\ [4pt]
    \midrule
    ours  & \textbf{0.9712} & \textbf{0.7558} \\ [4pt]
    \bottomrule
    \end{tabular}
\end{table}

\subsection{Intra Test Comparison}
We perform intra testing on four public datasets: FF++, Celeb-DF, DFDC, and WideDeepfake. Training and testing on the same dataset, this scheme aims to reveal the model's capacity of capturing the forged trace in deepfake videos. The binary accuracy is chosen as the evaluation metric and reported.

\noindent\textbf{Results on FF++.}
We first evaluate the model performance under different image quality settings in the FF++ dataset, where c23 stands for high quality and c40 stands for low quality. The comparison is illustrated in Table \ref{ffpp}. It is obvious that: (1) Our proposed method outperforms nearly all compared opponents on all settings except for F2F c23, which is slightly worse than S-MIL-T~\cite{li2020sharp}.
This proves the effectiveness of our proposed STIL block which is specially designed for the deepfake video detection task; (2) On the more challenging setting of c40, our method has a significantly higher performance over other counterparts by a considerable margin. Especially on the NT c40 setting, we achieved 91.78\% accuracy, exceeding 4.99\% than the state-of-the-art deepfake video detection method S-MIL-T. This is because, under this setting, the forgery technique mainly focuses on creating subtle facial artifacts which are further hidden by the low image quality, leading to a fairly genuine forgery effect. And our STIL block utilizes the temporal inconsistency and is still able to identify the forged samples, while other works encounter severe performance drop under this setting. Note that, TEI utilizes the inter-frame difference for temporal modeling and also achieves better performance than S-MIL-T in the NT c40 setting, proving the necessity of modeling temporal inconsistency. However, compared to our method, TEI is not specifically designed for this task and results in sub-optimal spatial-temporal inconsistency learning ability.


\begin{table}[htbp]
  \centering
  \caption{Ablation study on FaceForensics++ c40. Best results are highlighted.}
  \label{ablation}
    \begin{tabular}{lcccc}
    \multicolumn{5}{c}{(a) Study on module effectiveness.} \\ [2pt]
    \toprule
    Methods  & DF & F2F & FS & NT \\ [2pt]
    \midrule
    SIM      & 0.9536 & 0.8500  & 0.9286 & 0.8178 \\ [2pt]
    TIM      & 0.9464 & 0.8214 & 0.8821 & 0.7786 \\ [2pt]
    SIM+TIM  & 0.9821 & 0.9178 & 0.9678 & 0.8857 \\ [2pt]
    STIL    & \textbf{0.9821} & \textbf{0.9214} & \textbf{0.9714} & \textbf{0.9178} \\ [2pt]
    \bottomrule 
    \\
    \multicolumn{5}{c}{(b) Study on temporal difference operation.} \\ [2pt]
    \toprule
    Methods & DF & F2F & FS & NT \\ [2pt]
    \midrule
    S-diff & 0.9821 & 0.9214 & 0.9714 & 0.8750 \\ [2pt]
    H-diff & 0.9750 & 0.9143 & 0.9678 & 0.8893 \\ [2pt]
    W-diff & 0.9821 & \textbf{0.9286} & 0.9700 & 0.8786 \\ [2pt]
    H-diff + W-diff & \textbf{0.9821} & 0.9214 & \textbf{0.9714} & \textbf{0.9178} \\ [2pt]
    \bottomrule
    \\
    \multicolumn{5}{c}{(c) Study on information complementarity.} \\ [2pt]
    \toprule
    Methods & DF & F2F & FS & NT \\ [2pt]
    \midrule
    None                                 & 0.9821 & 0.9178 & 0.9678 & 0.8857\\ [2pt]
    $\mathcal{S} + \mathcal{T}$ & 0.9607 & 0.8964 & 0.9357 & 0.8857 \\ [2pt]
    $\mathcal{T}\rightarrow \mathcal{S}$ & 0.9607 & 0.8821 & 0.9357 & 0.8964 \\ [2pt]
    $\mathcal{S}\rightarrow \mathcal{T}$ & \textbf{0.9821} & \textbf{0.9214} & \textbf{0.9714} & \textbf{0.9178} \\ [2pt]
    \bottomrule
    \\
     \multicolumn{5}{c}{(d) Study on difference fusion.} \\ [2pt]
    \toprule
    Methods & DF & F2F & FS & NT \\ [2pt]
    \midrule
    $\sigma_{V+H}$           & 0.9821 & 0.9143 & 0.9678 & 0.9000\\ [2pt]
    $ \frac{1}{2}(\sigma_V + \sigma_H)$  & \textbf{0.9821} & \textbf{0.9214} & \textbf{0.9714} & \textbf{0.9178} \\ [2pt]
    \bottomrule
    \end{tabular}
\end{table}

\noindent\textbf{Results on Celeb-DF, DFDC and WideDeepfake datasets.} We also conduct comparisons on Celeb-DF, DFDC, and WideDeepfake datasets. From Table \ref{celeb}, we can easily observe that our proposed method consistently outperforms all the compared counterparts by a large margin, \emph{e.g.,} 0.34\% higher, 2.83\% higher, and 0.87\% higher accuracies than each best opponent on Celeb-DF, DFDC, and WideDeepfake datasets separately. It is interesting to notice that, the frame-based XN-avg shows performance superiority over many video-based methods on several datasets. This provides strong proof of the importance of spatial information in DeepFake video detection. And our method develops a two-stream modeling manner by taking both the spatial and temporal inconsistency into consideration,
which achieves the best performance.

\subsection{Cross Test Comparison}
To demonstrate the generalization of our method, we perform cross test by training and evaluating model on different datasets. Following the convention setting in~\cite{masi2020two}, we train our model on FF++ under c40 compression level and evaluate on Celeb-DF dataset. Table~\ref{cross-dataset} shows the results in terms of the Area Under Curve (AUC) metric. Again, our method exhibits significantly higher AUC than all listed opponents. We can also find that the frame-based methods have severe performance drop when transferred to the unseen Celeb-DF dataset than those video-based methods. This is reasonable because the spatial inconsistency feature varies along with the manipulation methods, and the frame-based methods are prone to overfitting to only seen forged patterns. In this setting, ignoring the temporal cues 
will inevitably lead to poor generalization capability. However, our STIL block establishes an optimal balance between spatial and temporal information, and the elaborately designed architecture further facilitates the generalization ability.

\section{ABLATION STUDY}
To systemically evaluate the model designs, we perform ablation studies on the FF++ dataset under compress rate $c40$ from four aspects and then give a complete visualization analysis.

\noindent\textbf{Study on module effectiveness.} We explore the effectiveness of each module in STIL block by simply removing them, \emph{i.e.}, we develop the following variants: 1) only SIM; 2) only TIM; 3) both SIM and TIM but without ISM; 4) complete STIL block with SIM + TIM + ISM. As illustrated in Table \ref{ablation} (a), we can find that only one single component from SIM or TIM decreases the accuracy a lot. We analyze that these modules have limited representation capacity to identify DeepFakes attacks. However, combining them greatly improves the performance as both spatial and temporal information is taken into consideration. Moreover, introducing ISM to passing information contributes to the best performance.

\noindent\textbf{Study on temporal difference operation.} We compare various forms of difference operation, including 1) spatial difference (S-diff), 2) horizontal difference (H-diff), 3) vertical difference (V-diff) and 4) dual path of difference (V-diff + H-diff). The spatial difference S-diff is adopted in TEI that does not slice along horizontal and vertical directions but on the whole feature map instead. As shown in Table \ref{ablation} (b), we can see that capturing temporal inconsistency from both horizontal and vertical directions obtains the best results, and directly from spatial difference gets worse results, especially on challenging NT attack type ($87.5\%$ vs $91.78\%$).
\begin{figure}[t!]
  \centering
  \setlength{\abovecaptionskip}{0.3cm}
  \includegraphics[width=\linewidth]{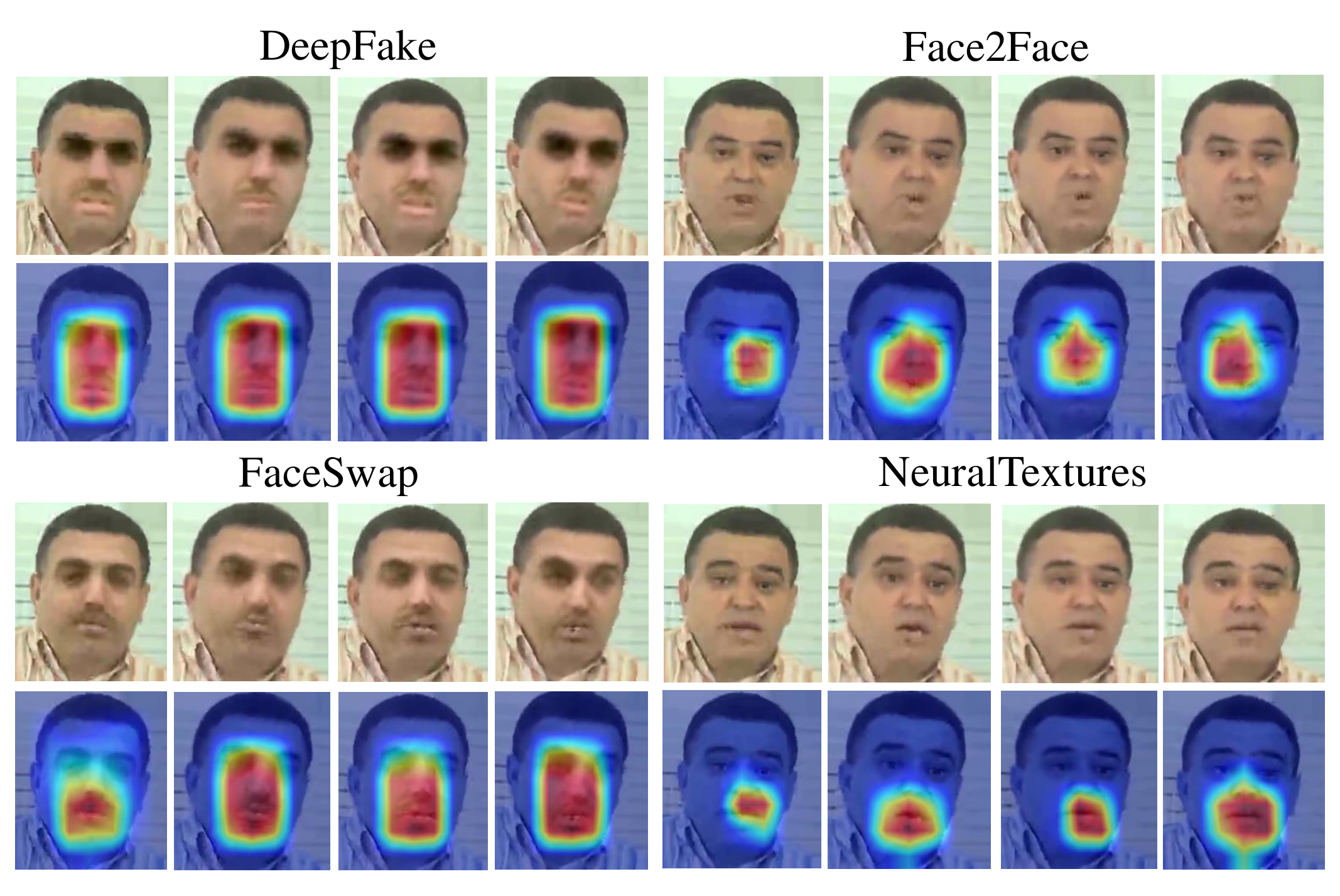}
  \caption{The visualization of the output from STIL block against four manipulations. The first and third row are fake sequences and the learned features are shown in the second and forth rows.}\label{differentheatmap}
\end{figure}

\begin{figure}[t!]
  \centering
  \setlength{\abovecaptionskip}{0.3cm}
  \includegraphics[width=8.5cm, height=4.5cm]{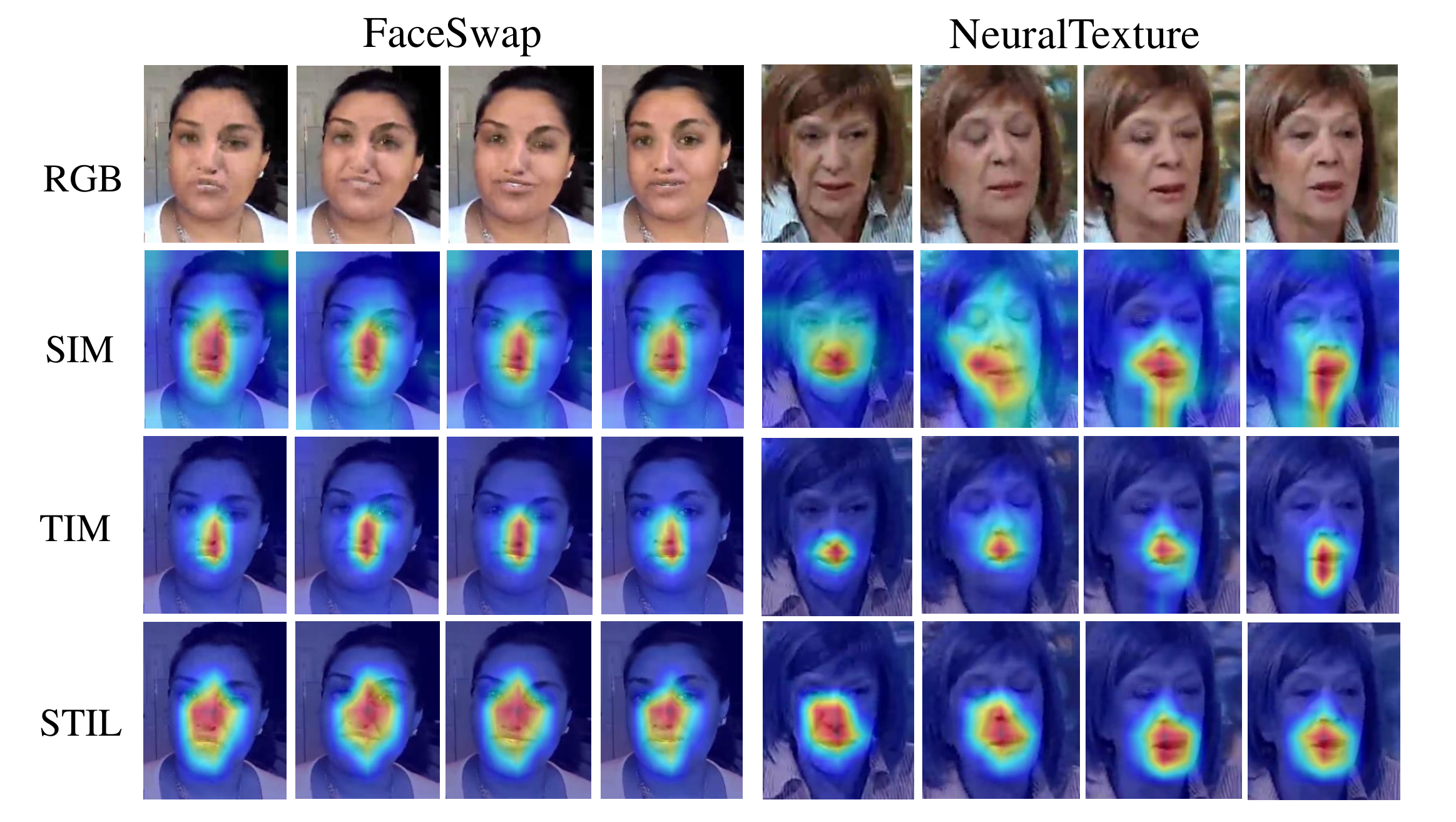}
  \caption{The visualization of attention maps from SIM, TIM and STIL.}\label{stbeforefinalheatmap}
  \Description{}
\end{figure}

\noindent\textbf{Study on information complementarity.} In terms of the complementarity between spatial and temporal information, we study three forms of information flow by building connections from one stream to the other: 1) unidirectional connection from SIM to TIM, denoted as $\mathcal{S}\rightarrow \mathcal{T}$; 2) unidirectional connection from TIM to SIM, denoted as $\mathcal{T}\rightarrow \mathcal{S}$; 3) bidirectional connections between SIM and TIM, denoted as $\mathcal{S} + \mathcal{T}$. The results are reported in Table \ref{ablation} (c). We observe that only $\mathcal{S} \rightarrow \mathcal{T}$ contributes to consistent performance gains. This is reasonable: spatial inconsistency can be considered as the temporal inconsistency with temporal kernel size $1$, leading to $Y_1$ and $Y_3$ with different temporal contexts. Otherwise, $Y_1$ will obtain similar temporal information with $Y_3$, which is redundant and thus has limited capacity.

\noindent\textbf{Study on difference fusion.} We conduct comparative studies to investigate the effects of different fusion strategies for combining temporal information from SIM and TIM, \emph{i.e.} simply averaging the two attention maps and summing temporal information from two directions before performing the sigmoid function. The results are summarized in the Table \ref{ablation} (d). As can be observed that simply averaging the attention maps from both horizontal and vertical directions gives consistently better results. We conjecture that the sum of temporal information from two directions before performing the sigmoid function may affect the inconsistency representations of a single direction, leading to inferior temporal modeling capacity.

\section{Visualization and Analysis}
In this section, we visualize the corresponding heatmaps with Grad-CAM \cite{selvaraju2017grad} to give an intuitive interpretation of how STIL works. As mentioned in \cite{rossler2019faceforensics++}, FF++ contains fake videos from four manipulation methods, \emph{i.e.}, DeepFake, FaceSwap, Face2Face and NeuralTextures. The DeepFake utilizes deep learning tools to replace faces and FaceSwap transfers the face region from a source video to a target video. Face2Face transfers the expressions of a source video to a target video while keeping the identity of the target person unchanged. NeuralTextures learns a neural texture of the target person from the original video data. When visually observed, DeepFake and FaceSwap manipulate nearly the whole facial region, Face2Face with fewer areas, and NeuralTextures mainly focuses on forging the mouth areas.

We visualize the output of the last STIL block against these four manipulation techniques in Fig. \ref{differentheatmap}. Both of the learned feature maps for DeepFake and FaceSwap focus on the whole face while maps for Face2Face and NeuralTextures localize most forged areas. This implies that our STIL block captures the characters of each manipulation method to some extent though it is trained with only video-level labels. 

\begin{figure}[t!]
  \centering
  \setlength{\abovecaptionskip}{0.3cm}
  \includegraphics[width=8.2cm, height=4.5cm]{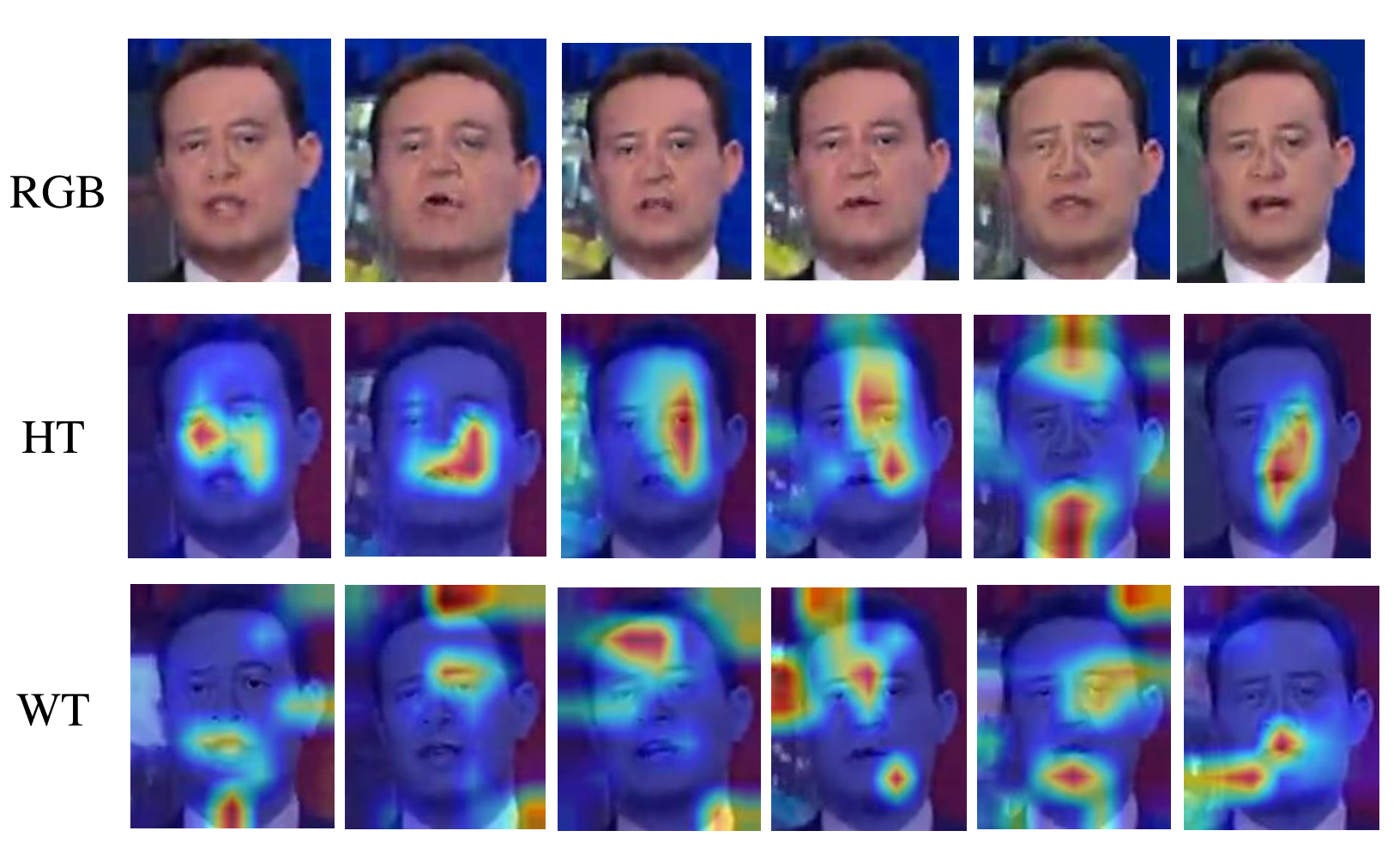}
  \caption{The visualization of vertical and horizontal attention maps from TIM.}\label{htwtheatmap} 
  \Description{A woman and a girl in white dresses sit in an open car.}
\end{figure}

To study what roles SIM and TIM play in STIL block and the effect brought by combining them, we visualize the corresponding heatmaps as illustrated in Fig \ref{stbeforefinalheatmap}. In the left sample of FaceSwap, both SIM and TIM locate the forged areas successfully but more or less not comprehensive enough. The combination of them does address this problem by focusing on the relatively challenging forged trace with a bigger highly activated map. In the right sample of NeuralTexture, although SIM focuses falsely on not forged areas in several frames, TIM utilizes the temporal information and their combination produces the final correct activation map. These two samples further exhibit the effectiveness of the two-stream design in the STIL block, and the fusion of spatial and temporal information leads to more comprehensive feature representations. 

In Fig.~\ref{htwtheatmap}, we visualize the $h-t$ and $w-t$ activation maps in video sequences and project them back to the original frames. We can find that in $h-t$ map, TIM tries to find the forged trace along the vertical direction, and in $w-t$ on the horizontal direction. Working as expected, TIM locates the forgery trace orthogonally along two most salient directions, which has been proven as the optimal solution in DeepFake video detection by the ablations.

\section{CONCLUSION}
In this paper, we formulate DeepFake video detection as a spatial and temporal inconsistency learning process and instantiate it into a STIL block, which effectively utilizes both the spatial and temporal information for a more comprehensive representation. To capture temporal inconsistency, we specially utilize the temporal difference over adjacent frames along with both horizontal and vertical directions within forged videos. 
Our STIL outperforms state-of-the-art methods on four widely used benchmarks. And in-depth ablation studies are conducted to investigate the effects of the block design. Moreover, extensive visualizations further demonstrate the eﬀectiveness of our method.

\begin{acks}
This research was supported in part by the National Natural Science Foundation of China (No. 61972157), National Key Research and Development Program of China (No. 2019YFC1521104), Shanghai Municipal Science and Technology Major Project (2021SHZDZX0102), and Zhejiang Lab (No. 2020NB0AB01).
\end{acks}

\bibliographystyle{ACM-Reference-Format}
\balance
\bibliography{sample-base}


\end{document}